\documentclass[11pt]{article}
\usepackage{main}
\usepackage{times}
\usepackage{url}
\usepackage{latexsym}
\usepackage{graphicx}
\usepackage{listings}
\usepackage{caption}
\usepackage{subcaption}
\usepackage{listings}
\usepackage{color}

\newcommand{\citep}[1]{\cite{#1}}
\newcommand{\citeyear}[1]{#1}
\newcommand{\citet}[1]{\cite{#1}}

\colingfinalcopy % Uncomment this line for the final submission

\date{}

\title{Detecting ESG topics using domain-specific language models and data augmentation approaches}

 \author{
 Tim Nugent \\
 Refinitiv Labs, London, UK \\
 {\tt tim.nugent@refinitiv.com} \\\And
 
 Nicole Stelea\thanks{\, This work was carried out while the second author was working at Refinitiv Labs.} \\\And
 Jochen L. Leidner$^{a,b}$\\
 Refinitiv Labs, London, UK\\
 University of Sheffield, Sheffield, UK\\
 {\tt  jochen.leidner@refinitiv.com} \\
 }

\begin{document}

\maketitle

\begin{abstract}
Despite recent advances in deep learning-based language modelling, many natural language processing (NLP) tasks in the financial domain remain challenging due to the paucity of appropriately labelled data. Other issues that can limit task performance are differences in word distribution between the general corpora -- typically used to pre-train language models -- and financial corpora, which often exhibit specialized language and symbology. Here, we investigate two approaches that may help to mitigate these issues. Firstly, we experiment with further language model pre-training using large amounts of in-domain data from business and financial news. We then apply augmentation approaches to increase the size of our dataset for model fine-tuning. We report our findings on an Environmental, Social and Governance (ESG) controversies dataset and demonstrate that both approaches are beneficial to accuracy in classification tasks.
\end{abstract}

\section{Introduction}

Recent advances in deep learning have led to state-of-the-art performance across a broad range of natural language processing (NLP) tasks. The introduction of distributed representations and models pre-trained using large corpora mean that traditional, task-specific feature engineering is no longer required, and neural approaches that employ such techniques typically outperform more conventional machine learning models.
% Deep learning, which initially referred to multi-layer perceptrons
% with more than one hidden (non-input/output) layer, has emerged as an
% effective toolbox of machine learning techniques. By now it includes methods
% for encoding text numerically (word embeddings) in ways to satisfy
% requirements like vector density, preservation of distributional
% context as well as language models pre-trained on billions of words and
% enabling substantial recent improvements on a range of natural language processing
% tasks.
%     Despite such advances in deep learning-based language modelling,
% many natural language processing (NLP) tasks in the financial domain remain
% challenging due to the paucity of appropriately labelled data. Natural
% data, in particular annotated data for supervised learning, remains scarce,
% and \emph{data augmentation} has been proposed as one method to mitigate this;0
% 0it originated in computer vision, and has been defined as a
% technique ``to improve model accuracy, generalisation, and to control overfitting'' \cite{Bloice-Stocker-Holzinger:2017:unpublished}.
Despite these advances, many NLP tasks in the financial domain remain challenging due to the paucity of appropriately labelled data. In particular, annotated data for supervised learning remains scarce, although \emph{data augmentation} has been proposed as one method to address this problem. Data augmentation originated in computer vision, and enables practitioners to increase the diversity of data for model training without collecting new examples. It enables both model accuracy and generalization to be improved, while controlling for over-fitting \cite{Bloice-Stocker-Holzinger:2017:unpublished}. Besides insufficient data, having data that is sampled from a different subject domain or textual genre can also limit performance: differences in word distributions between the corpora, as they are typically used to pre-train language models, and financial corpora, which often exhibit specialized language and notation conventions (\textit{e.g.}~company ticker symbology) can lead to a drop in task performance.\newline  

In this paper, we investigate two approaches to mitigate these issues: \emph{domain
adaptation} and \emph{data augmentation}. Firstly, we experiment with further language model
pre-training using large amounts of in-domain data from business and financial news. 
We then apply augmentation approaches to increase the size of our dataset for model fine-tuning. 
We report our findings on an Environmental, Social and Governance (ESG) controversies dataset, and demonstrate that both approaches are beneficial to performance in automatic classification tasks. ESG data can provide investors with company-specific metrics that attempt to quantify a company's environmental footprint, the degree to which they conduct their business in a socially responsible manner, and to what extent they are well-governed. Examples of ESG controversies include news stories about toxic waste spills (Environmental), human rights violations (Social), and corrupt CEOs (Governance)\footnote{Note that our definition is company- and investment -centric and differs from consumer/citizen-relevant controversies dealt with by \newcite{DoriHacohen-etal:2016:SIGIR} and \newcite{Choi-etal:2010:PAISI}, for example.}.
ESG controversies can be highly impactful on financial markets with recent estimates indicating they have wiped \$500bn off the value of US companies over the past five years\footnote{ \url{https://www.ft.com/content/3f1d44d9-094f-4700-989f-616e27c89599} (accessed 2020-06-30)}. Companies have already recognized that investors are taking ESG information
into account, which has also led to framing of stories in ESG terms more
favorably (``green'') than facts suggest, a practice known as greenwashing\footnote{
  \url{https://www.investopedia.com/terms/g/greenwashing.asp} (accessed 2020-06-30)
}. Recently, the United Nations defined the seventeen \emph{UN Sustainable Development Goals} \cite{lu2015policy} to promote global improvement of the human condition. Predictive models that can accurately classify news into ESG-related controversy categories, and also relate them to the UN SDGs, may be able to inform investment decision-making using news as actionable intelligence. This may be especially useful in an industry where ESG metrics are often published only annually or quarterly, providing investors with the ability to dynamically adjust their portfolios in response to emerging ESG news trends.
% \todox{ Space Permitting: add example of how that would look like }

%ESG controversy represent risks

%Artificial intelligence promises to enhance sustainable investing
%https://www.ft.com/content/7c40cdfc-b528-11e9-bec9-fdcab53d6959

%ESG/ML
%\cite{Allen2017}

\section{Related Work} \label{sec:related-work}

% We review the relevant machine learning and NLP literature as well as work in the application domain of responsible investment, and applications of machine learning or NLP to controversy detection or classification.

\subsection{Bidirectional transformer models (BERT)}

Bidirectional Encoder Representations from Transformers (BERT) is a neural language model capable of learning word representations from large volumes of unannotated text \cite{devlin-etal-2018-bert}. Unlike earlier approaches, BERT embeddings are highly contextual due to its deep and bidirectional (\textit{i.e.}~left-to-right and right-to-left) formulation. BERT pre-training uses a masked language model that learns to predicts words randomly masked from a sequence, and whether two sentences in a document naturally follow each other (next sentence prediction). It leverages a ``transformer'' encoder-decoder architecture which uses attention mechanisms to forward a more complete picture of the whole sequence to the decoder at once, rather than sequentially. Fine-tuned BERT models have been used to substantially improve performance in many downstream NLP tasks.
% that were randomly hidden (masked) during the training phase, in which
% its bidirectional ``transformer'' architecture  is exposed to documents. 
% hugely influential, accumulating several hundred citations
% between the publication of a preprint and the final paper, and it has been
% \\ \textbf{Domain-Specific Transformer Models.} 

\subsection{Domain-specific BERT variants} 
Domain-specific adaptation of BERT has
been conducted before, notably in the scientific domain and in
the financial sub-domain of securities filings:
    AllenAI's SciBERT \cite{beltagy2019scibert}, a pre-trained model
variant of BERT, was trained on large-scale labeled scientific data, and
leverages unsupervised pretraining on a large multi-domain corpus of
scientific publications. It was shown to improve performance in sequence tagging,
sentence classification and dependency parsing.
    In the biomedical domain, BioBERT \cite{lee2019biobert} was pre-trained on 
large-scale biomedical corpora. Again, this in-domain variant of BERT outperforms
vanilla BERT on biomedical named entity recognition (0.62\% F1 improvement),
biomedical relation extraction (2.80\% F1 improvement) and biomedical question 
answering (12.24\% MRR improvement).
    To support computational analysis of financial language, in particular SEC
quarterly and annual filings, \citet{Desola-Hanna-Nonis:2019:unpublished} present
FinBERT, a variant of BERT trained on decades of annual reports (Form 10-K,
1999-2019) and tested on quarterly reports (10-Qs).  FinBERT outperforms
BERT on several NLP performance tasks ($\approx{}30\%$ reduction in loss on next
sentence prediction; $\approx{}25\%$ reduction in loss in predicting masked words). The authors report a model trained on in-domain language
outperforms a model trained on Wikipedia and adapted on in-domain language. Another model, \emph{also} called FinBERT, targets the specific task of financial sentiment analysis as applied to a financial subset of the Thomson Reuters Corpus 2
(TRC2) %\footnote{
% TRC2 is available from NIST, cf.~\url{https://trec.nist.gov/data/reuters/reuters.html}
%}
dataset \cite{araci2019finbert}.

% \subsection{Deep learning with small datasets} 
% Neural architectures with many hidden layers
% are data-hungry, so recent research directions include how to make models work with smaller
% amounts of training data \cite{Barz-Denzler:2020:WACV}. Smaller dataset sizes are quite common,
% in particular when building systems serving professionals. \citet{Polson:1996:VIM} and \citet{Meng-vanDyk:1999:Biometrika} were among the first to introduce \emph{data augmentation} in the published literature, describing the automatic enrichment of a dataset by means of transforming the set of given training examples \cite{xie2019unsupervised}. One possible approach out of many for augmenting limited available data is \emph{back-translation}, the automatic translation from a given language into another and back into the source language, in order to create another variant of a given
% sentence \cite{xie2019unsupervised}.

\subsection{Responsible investing and ESG data}
Recently, there has been an increasing interest in incorporating ESG factors into investment decisions. \citet{In-Rook-Monk:2019:GlobEconRev} study how high quality ESG data can map onto investment decision-making processes. \citet{Friede-Busch-Bassen:2015:JSustFinInv} is a meta-study of over 2,200 studies linking ESG factors with financial performance. \citet{Schramade:2017:JApplCorpFin} provides a categorization of the 17 UN Sustainable Development Goals \cite{lu2015policy} into three ordinal classes indicating opportunity size.

\subsection{Automatic controversy classification} 

% <COMMENT>
% AFTER ACCEPTANCE, CHANGE TO: In ... we proposed a system ...
\newcite{Nugent-Leidner:2016:ICDM} propose a system for company risk profiling, which is related
to, but also different from controversy classification: while all controversies also represent a
risk, many but not all risks are controversial. Additionally, our method classifies documents,
while theirs classifies sentences. \newcite{DoriHacohen-etal:2016:SIGIR} and \newcite{Jang-etal:2016:CIKM} describe attempts to detect controversies in Wikipedia, whereas our approach is aimed at controversies reported in financial and business news. \newcite{Choi-etal:2010:PAISI} detect controversies in news, but the application is geopolitical security rather than financial investing. 
We are not aware of any previous attempts to train a deep learning model on a large financial news archive (\textit{i.e.}~several decades of data), and to the best of our knowledge, there is also no prior work on automatic ESG controversy classification.

\section{Methods}

\subsection{Financial corpus}

For pre-training, we start with the cased BERT-base model which consists of 12 layers, 12 attention heads, 768 hidden units, and a total of 110M parameters. This is trained using English Wikipedia and the BookCorpus \cite{zhu2015aligning} totalling approximately 3,300M words. Our financial corpus for further pre-training is the well-known \emph{Reuters News Archive (RNA)}, which consists of all Reuters articles published between 1996 and 2019\footnote{
  To obtain the data for replication, it can be licensed from Reuters at
  \url{https://www.reutersagency.com/en/products/archive/} (accessed 2020-06-30)}. We filter the corpus using metadata to ensure that only English language articles with Reuters topic codes that matched company news, corporate events, government finances or economic news were retained. Additionally, we excluded articles using topic codes and headline keywords that were news summaries, highlights, digests, and market round-ups. Such articles typically contain list or bullet points of unrelated news headlines which are unsuitable for next sentence prediction. Many financial news articles also contain ``structured'' data in the form of ASCII tables containing market data; we excluded these using a heuristic approach based on the fraction of non-alphabetical characters ($>$0.1). The resulting filtered corpus consists of 2.2M articles and 715M words.\newline

We converted this corpus into TensorFlow record format\footnote{\url{https://www.tensorflow.org/tutorials/load_data/tfrecord} (accessed 2020-06-30)}
for masked LM and next sentence prediction at sequence lengths of 128 and 512 tokens with a duplication factor of 10 and masked LM probability of 0.15. We performed sentence boundary disambiguation using spaCy\footnote{\url{https://spacy.io} (accessed 2020-06-30)}, but post-processed the results to correct errors around Reuters Instrument Codes (RICs) which are present in almost all articles. These symbols are used to identify financial instruments or indices and are made up of the security's ticker symbol, optionally followed by a period and exchange code based on the name of the associated stock exchange, all enclosed within angle brackets. For example, the RIC $<$IBM.N$>$ refers to IBM being traded on the New York Stock Exchange.

\subsection{Pre-training}

We performed pre-training using a maximum sequence length of 128 for 5M steps, using 50,000 warm-up steps, a learning rate of 1e-5 and a batch size of 256. We then additionally pre-train using a maximum sequence length of 512 for a further 1M steps, since long sequences are mostly needed to learn positional embeddings which can be learned relatively quickly. Pre-training was run on Google Cloud Tensor Processing Units (TPUs).

\subsection{ESG dataset}

\begin{table}
\centering
%\begin{small}
\begin{tabular}{p{6.05cm}p{5.75cm}p{1.25cm}}
\hline
ESG Controversy & UN SDG & Count \\ 
\hline
Accounting &  & 386 \\
Anti-Competition &  & 2945 \\
Business Ethics & 16 & 4672 \\
Consumer Complaints &  & 1386 \\
Customer Health \& Safety & 3 & 1479 \\
Diversity \& Opportunity & 5,9 & 904 \\
Employee Health \& Safety & 3 & 1427 \\
Environmental & 2,3,6,11,12,13,14,15 & 571 \\
General Shareholder Rights &  & 694 \\
Human Rights & 1,2,8 & 340 \\
Insider Dealings &  & 422 \\
Intellectual Property &  & 1875 \\
Management Compensation &  & 398 \\
Management Departures &  & 4082 \\
No Controversy &  & 5501 \\
Privacy &  & 791 \\
Public Health & 3,11 & 633 \\
Responsible Marketing & 1,3,4 & 1134 \\
Tax Fraud &  & 481 \\
Wages or Working Condition & 8 & 1484 \\
\hline
\end{tabular}
%\end{small}
\caption{Counts of each ESG controversy type, with mappings to the UN Sustainable Development Goals (SDGs) where available.}
\label{table:esg_dataset}
\end{table}

Our multi-class ESG dataset is a commercial offering provided by Refinitiv\footnote{\url{https://www.refinitiv.com/en/financial-data/company-data/esg-research-data} (accessed 2020-06-30)}, consisting of 31,605 news articles each annotated into one of 20 ESG controversy categories by analysts with backgrounds in finance and sustainable investing. Each article concerns an ESG controversy that a specific company is implicated in, with the entire dataset covering a total of 4137 companies over the time range 2002$-$2019. Included among these 20 is a negative class generated by sampling articles relating to companies in our dataset in the same date range whose headlines did not match a controversy article. There is significant class imbalance in the dataset, with only half of topics having $>$1,000 examples. Articles are processed by substituting company names identified using a named entity tagger with a placeholder. Additionally, we map 9 of these categories to 14 of the 17 UN Sustainable Development Goals (SDGs)\footnote{\url{https://sustainabledevelopment.un.org} (accessed 2020-06-30)}, resulting in a multi-label dataset consisting of 21,126 labels across 12,644 news articles. The ESG dataset along with the UN SDG mapping can be found in Table \ref{table:esg_dataset}.   

\subsection{Data augmentation}

We apply a back-translation approach for data augmentation \cite{sennrich2015improving,edunov2018understanding} to our training sets using the open-source Tensor2Tensor implementation \cite{tensor2tensor,xie2019unsupervised}.
Back-translating involves translating text in language A into another language B, before translating it back into language A, therefore creating an augmented example of the original text. This enables diverse paraphrases to be generated that still preserve the semantics of the original text, and can result in significant improvements to tasks such as question answering \cite{yu2018qanet}. We use the English-French and French-English models from WMT'14\footnote{\url{https://www.statmt.org/wmt14/translation-task.html} (accessed 2020-06-30)} to perform back-translation on each sentence in our training set.\newline  

\begin{table*}[t]
\centering
\begin{tabular}{p{1.5cm}p{13.5cm}}
\hline
Temp. & Paraphrase \\ 
\hline
\\[-1em]
-- & Human rights legal group says gold mining in Ghana rife with abuse, land grabs, pollution. \\ 
% 0.5 & A human rights legal group asserts that gold mining in Ghana is a source of violence, grazing and pollution. \\
0.6 & According to the human rights group, diamond mining in Ghana is subject to abuse, mountaineering and pollution. \\
0.7 & The human rights group asserts that gold mining in Ghana is a source of violence, friction and pollution. \\
0.8 & Ghana's human rights legislation had been consistent in affirming that the gold mines in Ghana were the target of abuse, land extraction and pollution. \\
0.9 & According to information provided by the human rights legal group, gold mining in Ghana is of paramount importance for fighting beautiful weather, property damage and pollution. \\
\hline
\end{tabular}
\caption{Examples of back-translated paraphrases generated using a range of softmax temperature settings. The first example is the original sentence from an article concerning mining company AngloGold Ashanti $<$ANGJ.J$>$.
}
\label{table:paraphrases}
\end{table*}

We controlled the diversity-validity trade-off of the generated paraphrases using a random sampling strategy controlled by a tunable softmax \emph{temperature}, ensuring diversity while preserving semantic meaning. A temperature setting of 0 results in identical paraphrases, while a setting of 1 results in diverse but incomprehensible paraphrases which risk altering the ground-truth label. We generated paraphrases using temperature settings in the range 0.6$-$0.9. Articles were split into sentences and paraphrases were generated for each sentence, before being recombined into articles. We generated 3 augmented articles at each temperature setting. Example paraphrases at these softmax temperatures are shown in Table \ref{table:paraphrases}.

\subsection{Fine-tuning}

% Document classification is achieved by appending an appropriate output layer onto the standard BERT architecture, before running the fine-tuning process. When used for inference on a news article, for examples:\newline

% \textit{Salmonella lawsuit filed against Wal-Mart. A lawsuit was filed today against Wal-Mart Stores, Inc., the company whose Greenwood, Indiana, store was the source of a Salmonella outbreak between May and August, 2006. The lawsuit was filed in Johnson County Superior Court on behalf of Ryan Merritt, a Greenwood resident whose son became violently ill and was hospitalized after consuming foods purchased at the Wal-Mart deli.}\newline

% A correct classification would results in the appropriate ESG category (Customer Health \& Safety), or UN SDG(s) (2 -- Good health and well-being), being predicted.\newline 

We fine-tune using cross-entropy loss after softmax activation for the multi-class task, and after sigmoidal activation for the multi-label task, respectively. Fine-tuning was run for 40 epochs with a dropout probability of 0.1, batch size of 64, maximum sequence lengths of 128, 256 and 512 tokens, and learning rates of 1e-5 and 1e-6. We used 30\% of the data for our test set, maintaining chronological ordering.

\subsection{Baseline methods}

For baseline comparison, we used a support vector machine (SVM) classifier with RBF kernel since SVMs are known to perform relatively well on small news-based datasets \cite{nugent2017comparison}. We use a one-vs-one scheme with TF-IDF features and sub-linear term frequency scaling. RBF gamma and regularization parameters are optimized using a grid search. We also used a hierarchical attention network (HAN) classifier \cite{yang-etal-2016-hierarchical} using GLoVe word embeddings \cite{Pennington14glove:global} which has been shown to outperform SVMs, CNNs, and LSTMs on document classification tasks.
\section{Results}
\subsection{ESG dataset fine-tuning}

Fine-tuning results for the  multi-class ESG dataset are shown in Table \ref{table:esg_results}. Here we can see that the baseline SVM approach (RBF gamma parameter 0.01, regularization parameter 1.0, F1=0.75) and the HAN model (GLoVE embeddings of dimension 200, F1=0.77) are both outperformed by all BERT models, each of which achieved highest performance using a maximum sequence length of 512 tokens, a learning rate of 1e-6 and a cased vocabulary. The relatively small improvement of the HAN over the SVM seems to demonstrate that the task is a challenging problem even for relatively recent deep learning architectures. BERT$_{BASE}$, which is pre-trained using general domain English Wikipedia and the BookCorpus only, improves on the HAN model by 0.01 in terms of absolute F-score (F1=0.78). BERT$_{RNA}$, our method further pre-trained on RNA, improves on BERT$_{BASE}$ performance by 0.04 (F1=0.82) which suggests that the additional pre-training using a financial domain corpus is effective in improving the performance of downstream classification tasks. The magnitude of improvement here is approximately in line with the performance gains we see elsewhere when additional domain-specific pre-training has been used \cite{lee2019biobert}. BERT$_{RNA-AUG}$, our method trained on RNA and fine-tuned using the augmented training set, achieves F1=0.84 when data is generated using a softmax temperature of 0.9, emphasising the importance of high diversity among training examples. This demonstrates that our back-translation approach is also effective, and that both approaches can be used in combination to achieve highest performance.\newline

Results for individual ESG controversies are shown in Table \ref{table:esg_category_results}. These show that the highest relative improvement in F1-score between BERT$_{RNA}$ and BERT$_{BASE}$ typically occurs in classes with amongst the lowest number of training examples, such as Accounting, Management Compensation, Public Health, Insider Dealings, and Human Rights controversies. We see an improvement in F1 score in 18 out of 20 classes, of which 16 are significant at the p$<$0.05 level, 2 at the p$<$0.01 level, and 13 at the p$<$0.001 level using McNemar’s test. Comparing BERT$_{RNA-AUG}$ with BERT$_{RNA}$, relative improvements are much smaller although we still see the largest gains in Accounting and Public Health controversies, and at least some improvement in 15 classes. 10 classes are significant at the p$<$0.05 level, and of these 3 at the p$<$0.01 level, and 5 at the p$<$0.001 level. Two classes -- Insider Dealings and Wages or Working Condition controversies -- actually see a slight reduction in performance, which suggests that the diversity-validity trade-off is slightly too high for these classes. 

\begin{table}
\centering
\begin{tabular}{p{5.6cm}p{1.6cm}p{1.6cm}p{1.6cm}p{0.3cm}}
\hline
Model & Precision & Recall & F1 \\ 
\hline
\\[-1em]
SVM$_{TF-IDF}$ & 0.76 & 0.75 & 0.75 \\
HAN$_{GLoVE-200}$ & 0.77 & 0.77 & 0.77 \\
BERT$_{BASE}$ & 0.79 & 0.78 & 0.78 \\
BERT$_{RNA}$ & 0.82 & 0.82 & 0.82 \\
BERT$_{RNA-AUG}$ & 0.84 & 0.84 & \textbf{0.84} \\
\\[-1em]
\hline
\end{tabular}
\caption{ESG dataset fine-tuning performance using. Our method is BERT$_{RNA}$, and BERT$_{RNA-AUG}$ uses data augmentation. HAN is the hierarhical attention network; best performance was achieved using GLoVe embeddings of dimension 200. Precision, recall and F1 scores are all weighted metrics accounting for class imbalance.}
\label{table:esg_results}
\end{table}

% %\begin{tabular}{lcccC{2.0cm}C{2.0cm}C{2.0cm}C{2.0cm}}
% \begin{table}[ht!]
% \centering
% \begin{tabular}{l{3.2cm}C{0.6cm}C{0.6cm}C{1.2cm}|}
% \hline
% ESG Controversy & BASE & RNA & RNA-AUG \\ 
% \hline
% \\[-1em]
% Accounting & 0.08  & 0.29 & 0.40 \\
% Anti-Competition & 0.79  & 0.84 & 0.84 \\
% Business Ethics & 0.65  & 0.70 & 0.71 \\
% Consumer Complaints & 0.48  & 0.53 & 0.59 \\
% Customer Health \& Safety & 0.70 & 0.74 & 0.76 \\
% Diversity \&  Opportunity & 0.85  & 0.86 & 0.88 \\
% Employee Health \& Safety & 0.81  & 0.85 & 0.87 \\
% Environmental & 0.59  & 0.66 & 0.70 \\
% General Shareholder Rights & 0.54  & 0.65 & 0.71 \\
% Human Rights & 0.53  & 0.67 & 0.70 \\
% Insider Dealings & 0.62  & 0.80 & 0.79 \\
% Intellectual Property & 0.92  & 0.92 & 0.94 \\
% Management Compensation & 0.57  & 0.76 & 0.82 \\
% Management Departures & 0.96  & 0.97 & 0.97 \\
% No Controversy & 0.99  & 0.99 & 1.00 \\ % Note that if you change None, you need to update the mention of it in the introduction!
% Privacy & 0.79  & 0.84 & 0.84 \\
% Public Health & 0.37  & 0.48 & 0.57 \\
% Responsible Marketing & 0.62  & 0.69 & 0.73 \\
% Tax Fraud & 0.58 & 0.71 & 0.73 \\
% Wages or Working Condition & 0.84  & 0.86 & 0.89 \\
% \\[-1em]
% \hline
% \end{tabular}
% \caption{
% ESG dataset fine-tuning F1 performance by BERT models for individual classes.
% BASE is the pre-trained using general domain English Wikipedia and the BookCorpus, RNA is our model further pre-trained on RNA, and RNA-AUG is our model further pre-trained on RNA and fine-tuned using the augmented training set.}
% \label{table:esg_category_results}
% \end{table}

\newcommand*{\SuperScriptSameStyle}[1]{%
  \ensuremath{%
    \mathchoice
      {{}^{\displaystyle #1}}%
      {{}^{\textstyle #1}}%
      {{}^{\scriptstyle #1}}%
      {{}^{\scriptscriptstyle #1}}%
  }%
}

\newcommand*{\oneS}{\SuperScriptSameStyle{*}}
\newcommand*{\twoS}{\SuperScriptSameStyle{**}}
\newcommand*{\threeS}{\SuperScriptSameStyle{*{*}*}}

%\begin{tabular}{lcccC{2.0cm}C{2.0cm}C{2.0cm}C{2.0cm}}
\begin{table}[ht!]
\centering
\begin{tabular}{p{8.0cm}p{2.0cm}p{2.0cm}p{2.0cm}}
\hline
ESG Controversy & BASE & RNA & RNA-AUG \\ 
\hline
Accounting & 0.08  & 0.29\threeS & 0.40\threeS \\
Anti-Competition & 0.79  & 0.84\threeS & 0.84 \\
Business Ethics & 0.65  & 0.70\threeS & 0.71 \\
Consumer Complaints & 0.48  & 0.53\oneS & 0.59\threeS \\
Customer Health \& Safety & 0.70 & 0.74 & 0.76 \\
Diversity \&  Opportunity & 0.85  & 0.86\twoS & 0.88 \\
Employee Health \& Safety & 0.81  & 0.85 & 0.87 \\
Environmental & 0.59  & 0.66\threeS & 0.70\twoS \\
General Shareholder Rights & 0.54  & 0.65\threeS & 0.71\oneS \\
Human Rights & 0.53  & 0.67\threeS & 0.70\twoS \\
Insider Dealings & 0.62  & 0.80\threeS & 0.79 \\
Intellectual Property & 0.92  & 0.92\threeS & 0.94\oneS \\
Management Compensation & 0.57  & 0.76\threeS & 0.82\threeS \\
Management Departures & 0.96  & 0.97 & 0.97 \\
No Controversy & 0.99  & 0.99 & 1.00 \\
Privacy & 0.79  & 0.84\threeS & 0.84 \\
Public Health & 0.37  & 0.48\threeS & 0.57\threeS \\
Responsible Marketing & 0.62  & 0.69\threeS & 0.73\threeS \\
Tax Fraud & 0.58 & 0.71\threeS & 0.73\twoS \\
Wages or Working Condition & 0.84  & 0.86\twoS & 0.89 \\
\hline
\end{tabular}
\caption{
ESG dataset fine-tuning F1 performance by BERT models for individual classes.
BASE is the pre-trained using general domain English Wikipedia and the BookCorpus, RNA is our model further pre-trained on RNA, and RNA-AUG is our model further pre-trained on RNA and fine-tuned using the augmented training set. Asterisks indicate statistical significance compared to the model in the previous column using McNemar's test at p-values of $<$ 0.05(\oneS), 0.01(\twoS) and 0.001(\threeS)}
\label{table:esg_category_results}
\end{table}

\subsection{Qualitative analysis}

We also performed a qualitative analysis by comparing the BERT$_{BASE}$ and BERT$_{RNA}$ confusion matrices and inspecting the off-diagonal values. Some of the more interesting misclassifications by BERT$_{BASE}$ that were corrected by BERT$_{RNA}$ were Accounting controversies and Management Departures controversies, for example:\newline

\textit{The Finnish company swung to a net loss of 1.6 billion euros in the first quarter, hit by falling sales and heavy restructuring charges. Nokia said Colin Giles, head of sales, would leave the firm in June, as it restructures the sales team. Nokia's first-quarter cellphone sales fell 24 percent from a year ago. The company said Giles was leaving to spend more time with his family and would not be replaced.}\newline 

Management Departures are often associated with negative earnings reports which, although are typically not controversial, often include similar references to financial losses as are found in actual Accounting controversies, for example:\newline

\textit{Three former Deutsche Bank employees have filed complaints with the U.S. securities regulators claiming the bank failed to recognize up to \$12 billion of unrealized losses during the financial crisis.}\newline

It appears that BERT$_{RNA}$ is able to learn to distinguish between such cases. Other examples include misclassifications between Public Health controversies and Customer Health \& Safety controversies; these are often hard to distinguish and occasionally overlap. The following example was correctly predicted as  Public Health controversy by BERT$_{RNA}$ and incorrectly predicted as Customer Health \& Safety controversy by BERT$_{BASE}$:

\textit{A wireless service provider has agreed to drop its lawsuit against the county over proposed wireless antennas in Hacienda Heights under one condition -- that it approve a permit that residents have vehemently opposed for almost four years. Between September 2008 and February 2009, the permit was approved twice by county officials and appealed both times by dozens of residents, including 13-year resident John Chen. ``Our residents and community are really, really concerned about health problems,'' said Chen, President Broadmoor Monaco Crest Homeowners Association.}
% Human Rights controversy and Environmental controversy misclassifications were also frequently corrected by BERT$_{RNA}$, for example:\newline  

% \textit{Environmental groups are accusing one of the world's largest paper companies of using land in China's south that was seized illegally from farmers by a local government, highlighting a problem that has fueled rural unrest and discontent across the country. The report released Friday by the Rights and Resources Initiative, a nonprofit group that promotes forest-use rights, and the Rural Development Institute does not accuse Finland-based paper manufacturer Stora Enso of forcing people from their land, but says a huge multinational should have known such abuses are common in China.}\newline  

% Here, BERT$_{RNA}$ was able to correctly identify that the article concerns forest-use rights rather than environmental issues, despite this being a difficult example.

\subsection{UN SDG dataset fine-tuning}

Fine-tuning results for the multi-label UN SDGs dataset are shown in Table \ref{table:sdg_results}. 
Similar to the ESG results, we are able to improve on BERT$_{BASE}$ (F1=0.75) using BERT$_{RNA}$ by 0.03 (F1=0.78) -- slightly less than the corresponding improvement in the ESG dataset. BERT$_{RNA-AUG}$ further improves over BERT$_{RNA}$ by 0.05 (F1=0.83), which is slightly more than in the ESG dataset. Again, best model performance was achieved using a maximum sequence length of 512 tokens, a learning rate of 1e-6, and a softmax temperature of 0.9 in the case of BERT$_{RNA-AUG}$. Due to the mapping in Table \ref{table:esg_dataset}, the broad Environmental ESG category corresponds exclusively to 5 different UN SDGs (6,12,13,14,15) and performance is highly correlated between them. If we treat these as a single class during validation, the relative performance of the three methods are unchanged with F1 scores 0.76, 0.79 and 0.85 for BERT$_{BASE}$, BERT$_{RNA}$ and BERT$_{RNA-AUG}$, respectively.  

\begin{table}
\centering
\begin{tabular}{p{3.6cm}p{1.6cm}p{1.6cm}p{1.6cm}p{1.6cm}}
\hline
Model & Loss & Precision & Recall & F1 \\ 
\hline
\\[-1em]
BERT$_{BASE}$ & 0.23 & 0.84 & 0.70 & 0.75 \\
BERT$_{RNA}$ & 0.16 & 0.85 & 0.73 & 0.78 \\
BERT$_{RNA-AUG}$ & 0.07 & 0.85 & 0.83 & \textbf{0.83} \\
\\[-1em]
\hline
\end{tabular}
\caption{UN SDGs dataset fine-tuning performance. Our method is BERT$_{RNA}$, and BERT$_{RNA-AUG}$ uses data augmentation.}
\label{table:sdg_results}
\end{table}

\subsection{Linguistic analysis of self-attention heads}

\begin{figure*}[htbp]
\centerline{\includegraphics[width=1.0\textwidth]{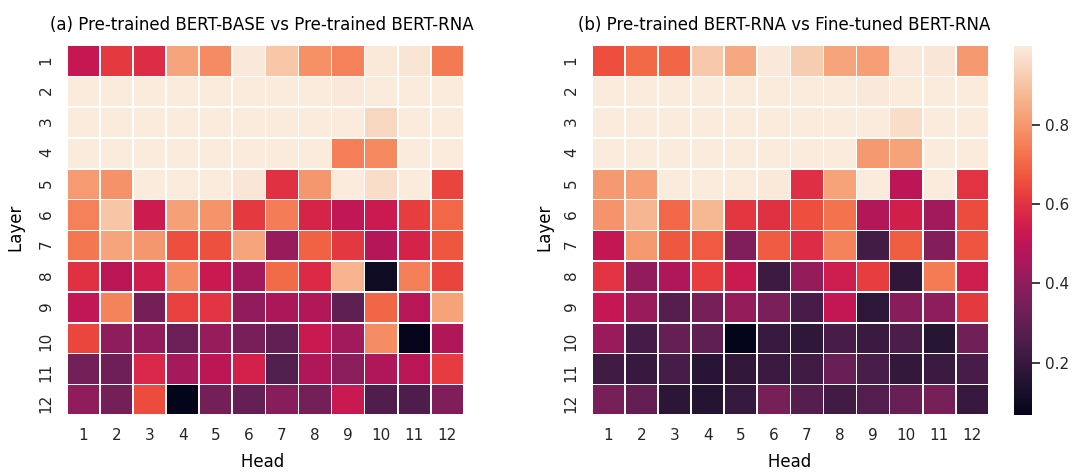}}
\caption{(a) Cosine similarity between pre-trained BERT$_{BASE}$ and pre-trained BERT$_{RNA}$, and (b) between pre-trained BERT$_{RNA}$ and fine-tuned BERT$_{RNA}$. Lighter cells indicate higher cosine similarity.}
\label{fig:attention}
\end{figure*}

To scrutinize linguistic behaviour, we also performed an analysis of BERT's self-attention weights for every head (12) in each layer (12) by passing the test set through our models and extracting these weights using bertviz \cite{vig2019transformervis}. This resulted in 144 LxL matrices, where L is the length of the input sequence plus the two special [CLS] and [SEP] tokens. As a first experiment, we compared these matrices between pre-trained BERT$_{BASE}$ and pre-trained BERT$_{RNA}$ models, in order to gauge the affect of additional domain-specific pre-training at the self-attention head level. Matrices were extracted and then flattened into arrays, allowing us to calculate cosine similarity between equivalent heads in the two models. Figure \ref{fig:attention}a shows the average cosine similarity scores across the entire test set. Results indicate that the latter layers (6-12) undergo the largest changes as a result of additional pre-training, while layers 2-5 are relatively unchanged in BERT$_{RNA}$. Perhaps surprisingly, the very first layer also sees significant changes in the majority of its heads. These patterns suggest that domain-specific pre-training contributes linguistic information to layers 1 and 6-12, while layers 2-5 seem to capture lower-level domain-agnostic knowledge. Comparing the self-attention patterns between pre-trained BERT$_{RNA}$ and fine-tuned BERT$_{RNA}$ in Figure \ref{fig:attention}b shows a similar pattern suggesting that ESG-specific information is in general learnt by the same heads. However, the last three layers in particular show substantially lower cosine similarity suggesting that they are responsible for learning task-specific features, a finding that has been observed previously \cite{kovaleva2019revealing}.\newline

In order to evaluate self-attention at the word level, we recombined words that had been split into multiple tokens by BERT's WordPiece tokenization. To calculate attention to a split word, we sum the attention weights over its tokens, and average the attention over its tokens when the word is attended to, maintaining the property that attention from a word always sums to 1. We then calculated the frequencies at which entities, dependency and part-of-speech (POS) tagged words are attended to by passing sentences from each test document through every model. Entities tend to be attended to most by a small number of heads between layer 8 and 11 in pre-trained BERT$_{BASE}$, while in pre-trained BERT$_{RNA}$ they are distribute in layers 1, 11 and 12, and in layers 1, 7 and 9 in fine-tuned BERT$_{RNA}$. Across all models, subjects are attended to mostly by a few heads in layers 6 and 7, with little difference between pre-trained and BERT$_{BASE}$ pre-trained BERT$_{RNA}$. In fine-tuned BERT$_{RNA}$ however, layers 9-12 show much higher attention to nominal subjects. A similar pattern can be observed for objects -- again being attended to by heads in layers 6-7 across all models with increased attention in layers 8-11 in fine-tuned BERT$_{RNA}$. While it was difficult to identify obvious differential POS tag patterns between models, fine-tuned BERT$_{RNA}$ showed increased attention to numerals (layers 6-10), verbs (layers 11-12), nouns (layer 11) and pronouns (layers 11-12), suggesting that these tags are especially important in ESG topic classification.

\section{Discussion and future work}

In this paper, we have introduced BERT$_{RNA}$ -- a domain-specific version of the BERT language model which has been further pre-trained using a 715M word corpus consisting of financial and business articles from Reuters News Archive. We have demonstrated that fine-tuning BERT$_{RNA}$ for two downstream classification task -- multi-class ESG controversy and multi-label UN SDGs detection -- can result in an improvement in performance compared to the general domain BERT$_{BASE}$ model. Furthermore, we have applied a back-translation approach to the limited dataset and have demonstrated that such techniques can further boost classification performance, and have also invested the linguistic behaviour of the different layers in both the pre-trained and fine-tuned models. Taken together, these results indicate that domain-specific language models and data augmentation can both help to mitigate the challenges associated with building machine learning models using small datasets. These approaches and the associated performance gains may be be especially important in the financial industry, which is highly adversarial, and could form the basis for alpha-generating NLP/NLU systems \cite{feuerriegel2016news}. To the best of our knowledge, this is the first use of a financial and business domain neural language model applied to ESG controversy prediction.\newline

While these results are encouraging, there are a number of areas where we believe further work could enable additional performance gains. One aspect which we did not address is the use of a domain-specific vocabulary. BERT's general domain vocabulary is constructed using WordPiece tokenization \cite{wu2016google}, which effectively deals with out-of-vocabulary issues by representing words using sub-word units (``wordpieces''). While financial and business vocabulary is closer to BERT's general domain vocabulary than for example, the life sciences, the use of acronyms and abbreviations (EBITDA, CAGR, IPO, P/E, \textit{etc.}) in news articles is extremely common. A domain-specific vocabulary may be of benefit here since many of these terms provide strong clues as to the subject of news articles. \newline

Another area of focus could be to extend the financial pre-training corpus beyond Reuters News to include other news sources, as well as additional sources of financial text such as SEC company filings and transcripts of earnings calls. Reuters News has specific style guidelines\footnote{\url{http://handbook.reuters.com} (accessed 2020-06-39)}, so we may be sacrificing stylistic diversity by using only a single source. The rich metadata provided in RNA does however allow us to filter the contents to ensure high quality and relevance, and this information is much more limited elsewhere. A closer inspection of the underlying mechanisms that contribute to the performance gains of domain-specific models may also yield useful insight. A more in-depth analysis of the self-attention head patterns could reveal whether certain linguistic features are more or less prevalent in the financial domain \cite{kovaleva2019revealing,clark-etal-2019-bert}, while extensions to the underlying transformer architecture may enable the processing of much longer financial texts (such as SEC company filings) than is currently possible \cite{kitaev2020reformer}. There are also more advanced methods for data augmentation available, such as the unsupervised data augmentation \cite{xie2019unsupervised}, which uses weighted supervised cross entropy and unsupervised consistency training loss, and combines well with BERT fine-tuning. Other loss functions such as cosine loss are also reported to benefit deep learning on small datasets \cite{Barz-Denzler:2020:WACV}. Although we have tested our model here on ESG and UN SDGs classification problems, we intend to extend this validation to a much broader range of financial NLP and NLU tasks. Examples include analysis of financial sentiment, named entity recognition (NER), relation extraction and question answering. We also intend to adapt the approaches described here to some of the many BERT variants that have emerged recently \cite{lan2019albert}. We aim to address these challenges in future work.

% UNCOMMENT AFTER REVIEW:
% \section*{Acknowledgments}
%
% We would like to thank Matt Harding for implementing a demonstration front-end to the classifiers
% described in this paper and to Amanda West for funding this research.

% \clearpage

\section*{Acknowledgments} We would like to thank Geoffrey Horrell and Amanda West for supporting this research, and the Refinitiv ESG team for discussions.

% \section*{Acknowledgments} We would like to thank our anonymous reviewers
% for valuable feedback, Geoffrey Horrell and Amanda West for supporting this
% research, and the Refinitiv ESG team for discussions.

\bibliographystyle{main}
\bibliography{main}

\end{document}